\documentclass{article} 
\usepackage{iclr2027_conference,times}

\usepackage{amsmath,amsfonts,bm}

\def\eqref#1{equation~\ref{#1}}

\def\1{\bm{1}}

\DeclareMathAlphabet{\mathsfit}{\encodingdefault}{\sfdefault}{m}{sl}
\SetMathAlphabet{\mathsfit}{bold}{\encodingdefault}{\sfdefault}{bx}{n}

\usepackage{hyperref}
\usepackage{url}
\usepackage{amsmath,amssymb}
\usepackage{graphicx}
\usepackage{booktabs}
\usepackage{multirow}
\usepackage{algorithm}
\usepackage{algorithmic}
\usepackage{xcolor}
\usepackage{enumitem}
\usepackage{subcaption}
\usepackage{float}

\title{\centering CAESAR: Clustering via Autonomous Embedding-Space Agglomerative Reorganization}

\author{Ilan Bacry \\
Arlequin AI \\
\texttt{ilan.bacry20@gmail.com}
\And
R\'emi Devaux \\
Arlequin AI \\
\texttt{remi@arlq.ai}
\And
Antoine Jardin \\
Arlequin AI \\
\texttt{antoine@arlq.ai}
}

\iclrfinalcopy
\begin{document}

\maketitle
\lhead{Preprint. Under review.}

\begin{abstract}

Clustering algorithms that operate on nearest-neighbor graphs, such as FINCH (First Integer Neighbor Clustering Hierarchy), depend heavily on the quality of the embedding space they are given. However, pretrained vision and language model embeddings are not optimized for this purpose. We propose CAESAR, a method that reorganizes a pretrained embedding space: a reorganization network is trained to pull mutual nearest neighbors together and push non-neighbors apart, yielding a reorganized embedding space substantially better suited to clustering. CAESAR offers a second major advantage: it never requires the number of clusters $K$. This matters because in realistic unsupervised settings, $K$ is typically unknown and discovering it is often part of the problem, yet most strong clustering methods take it as an input. We therefore design the entire CAESAR pipeline to infer $K$ rather than assume it is known. Empirically, reorganizing the embeddings consistently improves clustering over the raw space on both text and image datasets. Since the few deep clustering methods that also infer $K$ do not release their code, we complement controlled comparisons with methods that infer $K$ on the same embedding space by comparisons with strong deep clustering methods that are given the true $K$, giving them a substantial oracle advantage. Even so, CAESAR outperforms all of them on text, achieves the best results on the most challenging image benchmark and remains competitive on the others. Reorganizing pretrained embeddings thus emerges as a simple and powerful route to clustering realistic data, where classes overlap and the number of clusters is unknown.
\end{abstract}
\section{Introduction}
\label{sec:intro}

Clustering is central to exploratory data analysis, yet most modern pipelines rely on embeddings from pretrained encoders that were never trained with clustering as an explicit objective. The so-obtained \emph{pretrained embedding space} is typically built for optimizing objectives such as next-token prediction or contrastive alignment across augmentations. These objectives do not guarantee that the pretrained embedding space exposes a neighborhood structure that clustering algorithms can exploit effectively. This is particularly acute for large language and vision-language models, whose embeddings are known to be anisotropic~\citep{mu2017all, gao2019representation, ethayarajh2019contextual} and optimized for generation or retrieval rather than for grouping semantically similar points together.

A second difficulty in any realistic clustering setting is that the number of clusters is unknown. In genuinely unsupervised applications, such as exploring an unlabeled corpus or discovering categories in new data, one does not know in advance how many groups exist. Determining this number is one of the main goals of the analysis. Most strong clustering methods sidestep this by requiring the number of clusters as an input, which quietly assumes away the hardest part of the problem. Graph-based algorithms such as FINCH~\citep{sarfraz2019efficient}, by contrast, build their partitions directly from first-neighbor relations and do not need the number of clusters in advance. This makes them a natural choice in this setting. The price of this flexibility is a strong dependence on the neighborhood structure of the pretrained embedding space, a property that, as discussed above, pretrained encoders do not guarantee. FINCH has no global objective to compensate for noisy or inconsistent neighborhoods, so a few spurious first-neighbor links can propagate through the entire hierarchy. Centroid-based methods such as $k$-means are more robust in this respect, but only because they are given the number of clusters in advance. The two problems thus compound: the methods best suited to an unknown number of clusters are also those most vulnerable to embeddings not designed for clustering.

We propose CAESAR (Clustering via Autonomous Embedding-Space Agglomerative Reorganization), a method that addresses both problems jointly. CAESAR takes the pretrained embedding space and reorganizes it to make it more amenable to graph-based clustering. It does so without retraining or modifying the encoder and without ever requiring the number of clusters. To address the first problem, we train a lightweight \emph{reorganization network} that maps the pretrained embedding space to a new space, the \emph{reorganized embedding space}. The reorganization network is trained to pull mutual nearest neighbors in the pretrained embedding space together and push non-neighbors apart. A memory bank allows each point to be compared against the full dataset rather than only the current batch. The reorganized embedding space is then clustered with FINCH. For the second problem, every stage of the pipeline is designed to operate without knowing the number of clusters: the neighborhood size is estimated from the geometry of the pretrained embedding space, the balance between the pull and push terms is set to a single value shared across all datasets and the final partition is chosen from the FINCH hierarchy by an unsupervised criterion, never by matching a target number of clusters.

We evaluate CAESAR on text and image datasets with varying levels of class separability. CAESAR consistently improves over FINCH applied directly to the pretrained embedding space. 
Since existing deep clustering methods that infer $K$ are not reproducible from publicly available code, we complement our controlled comparisons with published results from recent deep clustering methods that use different encoders and are given the true number of clusters.
Despite this and without ever being given the number of clusters, CAESAR outperforms every competing method on text and remains competitive on images, sometimes surpassing methods that know the true number of clusters. 
CAESAR is thus particularly well suited to the realistic setting of overlapping classes and an unknown number of clusters.

Our contributions are as follows:
\begin{itemize}[leftmargin=*]
    \item We propose a pull--push objective over mutual $k$-nearest neighbors, computed against a memory bank of the full dataset, to reorganize the pretrained embedding space of frozen LLM/VLM encoders for graph-based clustering.
    \item The entire pipeline operates without knowing the number of clusters: the neighborhood size is estimated from the data, the pull--push balance is shared across all datasets and the final partition is selected from the FINCH hierarchy by an unsupervised criterion.
    \item Across text and image datasets and without ever being given the number of clusters, CAESAR outperforms all competing methods on text and remains competitive on images, where it obtains the best NMI on the most entangled dataset---ahead even of baselines given the true number of clusters.
\end{itemize}

\section{Related Work}
\label{sec:related}

\paragraph{Deep clustering.}
A large body of work jointly learns representations and cluster assignments. DEC~\citep{xie2016unsupervised} and IDEC~\citep{guo2017improved} refine an autoencoder latent space toward a cluster-friendly structure by minimizing a KL divergence to a sharpened assignment distribution. DeepCluster~\citep{caron2018deep} alternates between clustering features and using the assignments as pseudo-labels. SCAN~\citep{van2020scan} first mines nearest neighbors with self-supervised pretraining, then trains a classification head that encourages consistent predictions among neighbors. CC~\citep{li2021contrastive} performs contrastive learning simultaneously at both the instance and the cluster level. SPICE~\citep{niu2022spice} splits the clustering network into a feature model and a clustering head and trains them in successive stages with semantics-aware pseudo-labels. 
These methods learn their representations end-to-end from raw inputs, typically rely on image augmentations and require the number of clusters $K$ as an architectural parameter (e.g., the width of the cluster head). In contrast, we operate on the pretrained embedding space of frozen foundation models, use no augmentations and never require $K$ at any stage.

\paragraph{Clustering with external textual guidance.}
A recent line of work improves image clustering by exploiting textual knowledge from vision-language models such as CLIP~\citep{radford2021learning}. SIC~\citep{cai2023semantic} maps images into a semantic space built from WordNet nouns, derives pseudo-labels from image--text similarities and enforces consistency in both the image and the semantic space. TAC~\citep{li2023image} selects the WordNet nouns that best discriminate the images, then mutually distills neighborhood information between the text and image modalities. Both methods keep the pretrained encoder frozen but rely on an external vocabulary and require $K$. In contrast, CAESAR uses no external textual supervision and never requires $K$.

\paragraph{Nearest-neighbor supervision for representation learning.}
Closest to our objective are methods that use nearest neighbors as positives in a contrastive loss. NNCLR~\citep{dwibedi2021little} retrieves the nearest neighbor of each view from a support queue and treats it as a positive, showing that cross-sample positives provide richer semantic variation than augmentations alone. NNM~\citep{dang2021nearest} matches samples to their nearest neighbors at both the batch (local) and dataset (global) level for deep clustering. Our loss follows this general pull--push principle, with three differences. First, we rely on \emph{mutual} $k$-nearest neighbors -- pairs that select each other -- which prunes asymmetric, boundary-crossing pairs from the positive set. Second, our memory bank stores the embeddings of the \emph{entire} dataset in the reorganized embedding space, so the repulsion term is computed against all points rather than a sampled queue. Third, whereas NNM and NNCLR learn a full encoder from images with augmentations, we train only a reorganization network on the pretrained embedding space of frozen LLM/VLM encoders, which makes the approach applicable to any modality for which a pretrained encoder exists.

\paragraph{Attraction--repulsion embedding methods.}
Our objective is also closely related to the attraction--repulsion principle at the heart of neighbor-graph embedding methods for dimensionality reduction and visualization. t-SNE~\citep{van2008visualizing} and UMAP~\citep{mcinnes2018umap} build a $k$-nearest-neighbor graph in the input space and optimize a low-dimensional layout in which neighboring points attract while non-neighbors repel. More recently, DiRE~\citep{kolpakov2025dimensionality} makes this attraction--repulsion trade-off explicit. CAESAR differs in intent and in setting: rather than projecting to two or three dimensions for visualization, we learn a reorganization into a moderate-dimensional space whose sole purpose is to improve downstream graph-based clustering. Moreover, our repulsion is computed against a full memory bank rather than through the sampling or negative-approximation schemes these methods use for scalability.

\paragraph{Clustering without a predefined number of clusters.}
FINCH~\citep{sarfraz2019efficient}, which serves as the clustering engine of our pipeline, builds a hierarchy of partitions from first-neighbor relations alone, without distance thresholds or a target number of clusters. Density-based methods such as DBSCAN and HDBSCAN~\citep{ester1996density,campello2013density} also avoid specifying $K$ but replace it with density hyperparameters that are notoriously hard to tune in high dimensions. 

EVoC~\citep{bot2025persistent}, designed specifically for embedding vectors, combines a UMAP-like node embedding with HDBSCAN-style density-based clustering and selects stable partitions through persistence analysis, also without a target number of clusters; unlike CAESAR, it optimizes a non-parametric low-dimensional layout rather than training a reorganization network. 
Our contribution is complementary to this line of work: rather than proposing a new $K$-free clusterer, we show that a learned reorganization of the embedding space substantially improves the partitions that such graph-based methods discover, while preserving their $K$-free property end to end -- the final partition is selected from the FINCH hierarchy with an internal validity criterion (silhouette;~\citealp{rousseeuw1987silhouettes}) rather than by matching a known $K$.

\paragraph{Geometry of pretrained embeddings.}
Pretrained language-model embeddings are known to be anisotropic~\citep{mu2017all,gao2019representation,ethayarajh2019contextual}, which motivates global post-processing such as BERT-flow~\citep{li2020sentence}, whitening or contrastive fine-tuning~\citep{gao2021simcse}. These corrections are task-agnostic. Graph-based clustering instead requires a clean \emph{local neighborhood structure}, which CAESAR targets directly by preserving mutual-neighbor relations.

\section{Method}
\label{sec:method}

\subsection{Overview}
\label{sec:overview}
\begin{figure}[t]
    \centering
    \includegraphics[width=\textwidth]{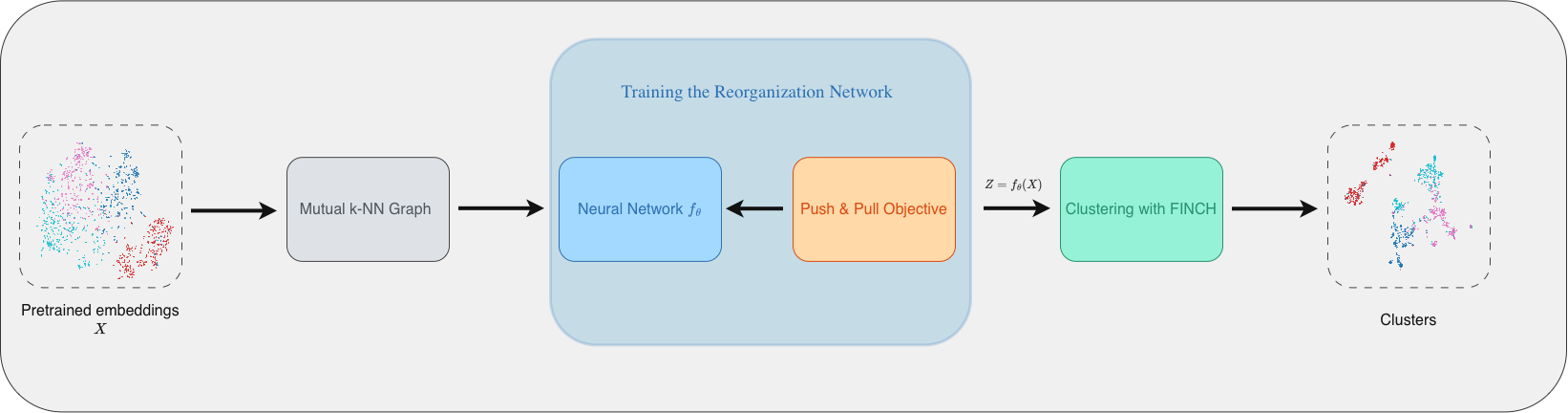}
    \caption{Overview of CAESAR. Starting from a pretrained embedding space $X$, we build a mutual $k$-NN graph (Section~\ref{sec:graph}) and train a reorganization network $f_\theta$ (Section~\ref{sec:transform_net}) with a pull--push objective (Section~\ref{sec:objective}). The resulting embeddings $Z = f_\theta(X)$ are partitioned by FINCH (Section~\ref{sec:downstream}). No stage of the pipeline uses the number of clusters.}
    \label{fig:overview}
\end{figure}

Our approach starts from a pretrained embedding space $X \in \mathbb{R}^{N \times d}$ produced by a foundation model and learns a reorganization network $f_\theta : \mathbb{R}^d \rightarrow \mathbb{R}^{d'}$ (Section~\ref{sec:transform_net}) that maps the data into a reorganized embedding space $Z = f_\theta(X)$, with $d' < d$. Figure~\ref{fig:overview} gives an overview of the full pipeline. Rather than learning representations from scratch, we aim to reshape the geometry of the pretrained embedding space while preserving its semantic structure. To this end, we build a neighborhood graph on $X$ (Section~\ref{sec:graph}) and train $f_\theta$ to preserve and sharpen these local relations in the reorganized embedding space: neighbors are pulled together, non-neighbors are pushed apart (Section~\ref{sec:objective}). This is done with a two-term loss (pull term and push term). The neighborhood size is estimated from the geometry of $X$. The embeddings $Z$ are then passed to FINCH and the final partition is chosen from its hierarchy by an unsupervised criterion (Section~\ref{sec:downstream}), never by matching a target number of clusters. This makes the framework directly applicable to realistic unsupervised settings, where the number of clusters is unknown.

Note that every choice in the pipeline is made without knowing the number of clusters.

\subsection{The Mutual $k$-NN Graph}
\label{sec:graph}
Before training $f_\theta$, we define the structure it must preserve: a neighborhood graph built, once for all, from the pretrained embedding space $X$. For each point $i$, we take its $k$ nearest neighbors $\mathcal{N}_k(i)$ under cosine similarity.

However, directed neighbors are unreliable for our objective. A point $j$ can be among the nearest neighbors of $i$ without $i$ being among those of $j$, especially in dense regions. We therefore keep only \emph{mutual} neighbors, $\mathcal{M}_k(i) = \{ j \in \mathcal{N}_k(i) : i \in \mathcal{N}_k(j) \}$. If $\mathcal{M}_k(i)$ is empty, we fall back to the single closest neighbor so that every point keeps at least one anchor. If fewer than $k$ mutual neighbors are found, $\mathcal{M}_k(i)$ is padded to $k$ entries by repeating its last element for batching.

\paragraph{Choice of $k$.}
Rather than fixing $k$ by hand, we estimate it from the geometry of the pretrained embedding space. On a random subsample of $X$, we compute the mean distance $\bar d(r)$ to the $r$-th nearest neighbor for $r = 1, \dots, k_{\max}$. This curve describes how fast the local neighborhood expands with rank. Its elbow marks the rank beyond which additional neighbors become markedly less informative. We locate this elbow as the rank of maximum discrete curvature,
\begin{equation}
    k = \arg\max_{r} \; \Delta^2 \bar d(r), \qquad \Delta^2 \bar d(r) = \bar d(r{+}2) - 2\bar d(r{+}1) + \bar d(r).
\end{equation}
The neighborhood size is thus set by the data geometry alone and is never tuned against clustering performance.

The graph $\mathcal{M}_k$ is computed once and kept fixed throughout training: it encodes the local structure of $X$ that $f_\theta$ must preserve and serves as the target of the objective defined in Section~\ref{sec:objective}.

\subsection{The Reorganization Network $f_\theta$}
\label{sec:transform_net}

The reorganization network $f_\theta$ is a lightweight neural network. Unlike a full encoder trained from raw data, it operates directly on the pretrained embedding space $X \in \mathbb{R}^{N \times d}$: its input dimension $d$ is set by the chosen pretrained model rather than by the raw modality. It outputs embeddings $Z \in \mathbb{R}^{N \times d'}$ in the reorganized embedding space, with $d' = 256$ across all datasets and modalities. Operating on an already informative space rather than learning an encoder from scratch keeps the number of trainable parameters small and restricts $f_\theta$ to reorganizing the geometry of $X$ instead of rediscovering the semantic content the foundation model has already captured.

The network is a three-layer MLP, $d \rightarrow 1024 \rightarrow 512 \rightarrow d' = 256$, with the same hidden sizes for all pretrained models, regardless of their input dimension $d$. Each hidden layer is followed by a SELU activation~\citep{klambauer2017self} and the output layer is linear.

\subsection{Training $f_\theta$ with the Pull--Push Objective}
\label{sec:objective}
We train $f_\theta$ with an objective designed to preserve the neighborhoods defined by $\mathcal{M}_k$, which brings mutual neighbors together in the reorganized embedding space and keeps non-neighbors apart. Let $z_i = f_\theta(x_i)$ be the embedding of point $i$ in the reorganized embedding space and $\hat z_i = z_i / \lVert z_i \rVert_2$ its $\ell_2$-normalized version, so that $\hat z_i^\top \hat z_j$ is the cosine similarity between $i$ and $j$.

Since a mini-batch $\mathcal{B}$ covers only a small fraction of the dataset, most neighbors of a point $i$ are not in $\mathcal{B}$ and restricting the objective to pairs within the batch would ignore most of $\mathcal{M}_k$. We therefore maintain a memory bank $\mathrm{Bank} \in \mathbb{R}^{N \times d'}$~\citep{wu2018unsupervised}, whose $j$-th row $\hat b_j$ stores the most recent $\ell_2$-normalized embedding $\hat z_j$ of point $j$ in the reorganized embedding space; after each optimization step, the rows of the points in $\mathcal{B}$ are updated. Each point in the batch can thus be compared to all its neighbors and to all other points, with a single forward pass of $f_\theta$ per batch.

The objective has two terms. Let $n_{i,1}, \dots, n_{i,k}$ denote the elements of $\mathcal{M}_k(i)$ ordered by increasing distance to $i$ in the pretrained embedding space. The pull term encourages each point $i$ to be similar to its neighbors:
\begin{equation}
    \mathcal{L}_{\text{pull}} = \frac{1}{|\mathcal{B}|} \sum_{i \in \mathcal{B}} \sum_{r=1}^{k} w_r \left(1 - \hat z_i^\top \hat b_{n_{i,r}}\right),
\end{equation}
where the weights $w_r \propto 1/r$, normalized to sum to one, give more influence to the nearest neighbors, which are more reliable positives than farther ones.

The push term discourages $i$ from becoming similar to points that are \emph{not} its neighbors, over the entire bank rather than only the current batch:
\begin{equation}
    \mathcal{L}_{\text{push}} = \frac{\lambda}{|\mathcal{B}| \, N} \sum_{i \in \mathcal{B}} \sum_{j=1}^{N} \left(1 - m_{i,j}\right) \max\left(\hat z_i^\top \hat b_j,\; 0\right),
\end{equation}
where $m_{i,j} = 1$ if $j = i$ or $j \in \mathcal{M}_k(i)$ and $0$ otherwise and $\lambda$ balances push against pull (Section~\ref{sec:hyper}). The full objective is $\mathcal{L} = \mathcal{L}_{\text{pull}} + \mathcal{L}_{\text{push}}$.

The clamp at zero penalizes a non-neighbor pair only while its similarity is positive; once their similarity is zero or negative, it no longer contributes to the gradient. Without the clamp, the objective would push every non-neighbor pair toward maximal dissimilarity. This is geometrically impossible: the average pairwise cosine similarity of $N$ unit vectors is at least $-1/(N-1)$. It would also let the push term, which sums over far more pairs than the pull term, dominate the gradient.

\begin{figure}[t]
    \centering
    \begin{subfigure}[b]{0.4\textwidth}
        \centering
        \includegraphics[width=\textwidth]{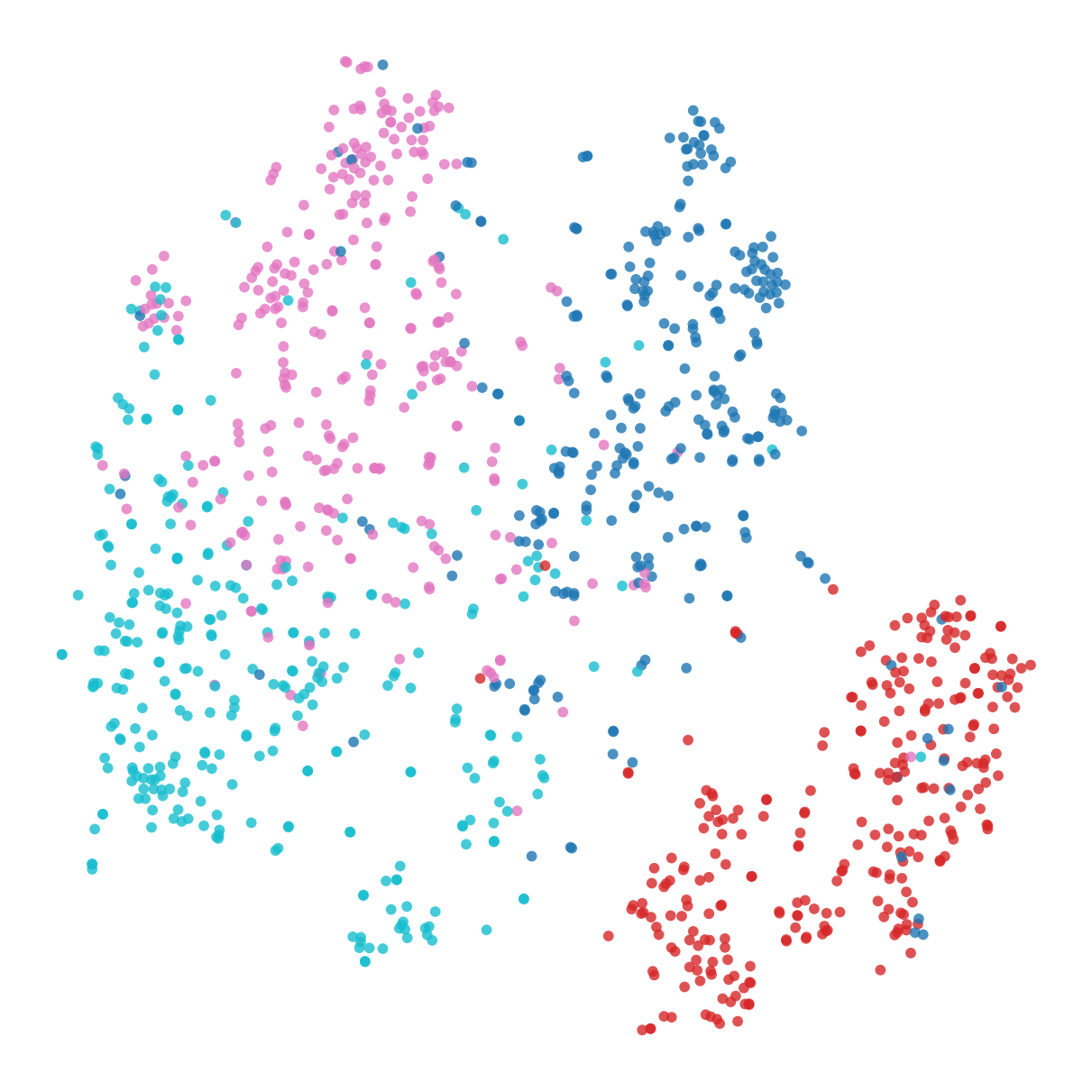}
        \caption{Pretrained embedding space}
        \label{fig:tsne_before}
    \end{subfigure}
    \hspace{0.05\textwidth}
    \begin{subfigure}[b]{0.4\textwidth}
        \centering
        \includegraphics[width=\textwidth]{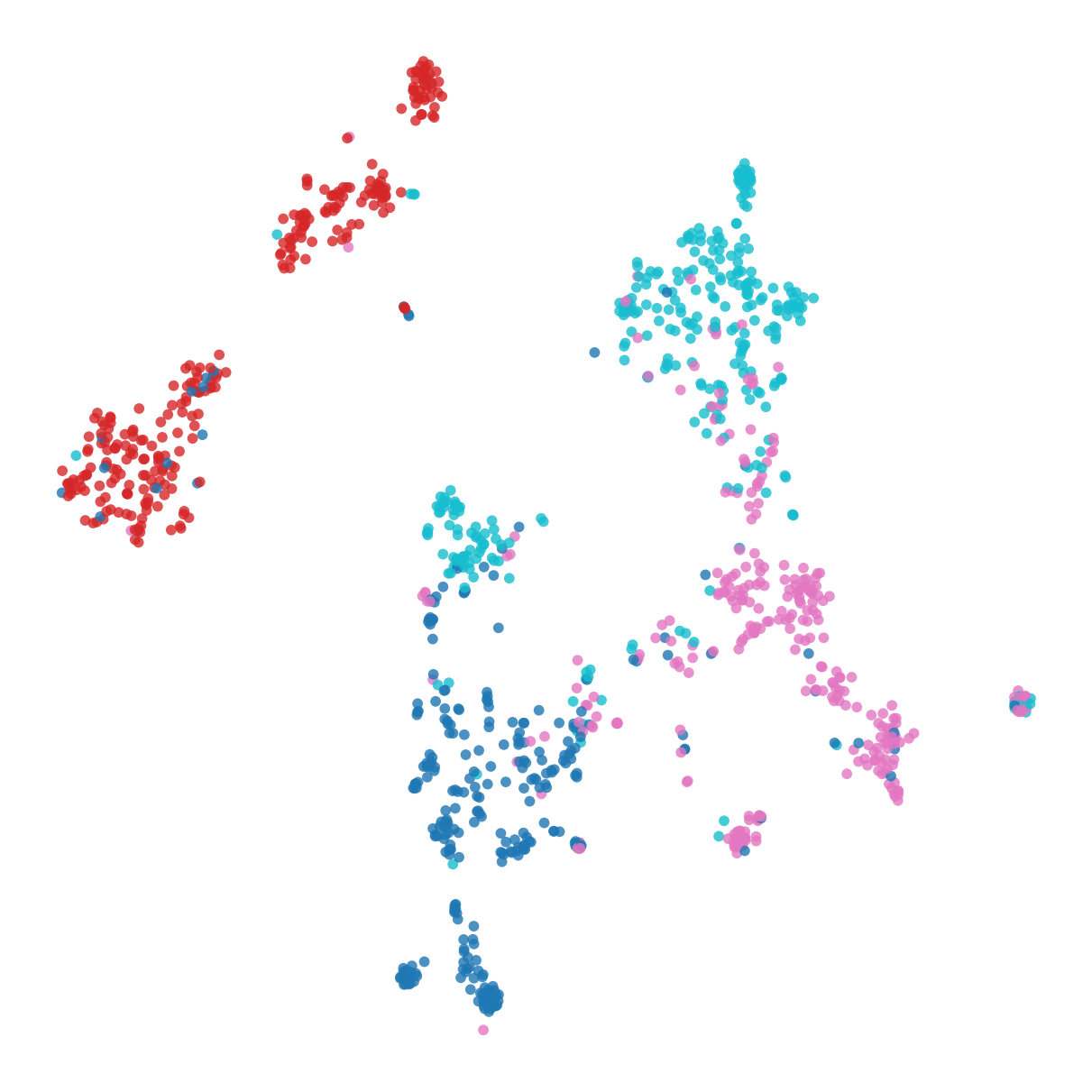}
        \caption{Reorganized embedding space}
        \label{fig:tsne_after}
    \end{subfigure}
    \caption{Effect of the pull--push objective on AG News, visualized with t-SNE~\citep{van2008visualizing}. Colors denote the four ground-truth classes and are used only for visualization, never during training. In the pretrained embedding space (left) the classes overlap; in the reorganized embedding space (right) they form compact, well-separated groups.}
    \label{fig:tsne}
\end{figure}

Figure~\ref{fig:tsne} illustrates the effect of this objective on AG News: in the pretrained embedding space the four classes overlap, while in the reorganized embedding space they form compact, well-separated groups.

\subsection{Clustering with FINCH}
\label{sec:downstream}
The embeddings $Z = f_\theta(X)$ in the reorganized embedding space are $\ell_2$-normalized and clustered with FINCH~\citep{sarfraz2019efficient}, which links each point to its first nearest neighbor and merges the resulting connected components recursively. It therefore requires no number of clusters in advance, consistent with the rest of our pipeline. Moreover, since FINCH relies on the same local neighbor relations that our objective reinforces, it directly reflects whether $f_\theta$ has achieved its intended effect.

FINCH returns a hierarchy of partitions ranging from fine to coarse and a single level must be selected. To avoid using the number of clusters, we select this level with an internal validity criterion: we compute the silhouette score~\citep{rousseeuw1987silhouettes} of each level in the reorganized embedding space and keep the level that maximizes it. As a baseline given the true number of clusters, we also report FINCH on the pretrained embedding space with oracle level selection: FINCH builds its hierarchy without $K$ and we then use the true $K$ to select the level whose number of clusters is closest to it.
\section{Experiments}
\label{sec:experiments}

\subsection{Experimental Setup}
\label{sec:exp_setup}

\paragraph{Datasets.}
We evaluate CAESAR on four datasets spanning two modalities. For text, we use AG News~\citep{zhang2015character} ($4$ topic classes). For images, we use CIFAR-10~\citep{krizhevsky2009learning} ($10$ classes), CIFAR-20 (the $20$ superclasses of CIFAR-100~\citep{krizhevsky2009learning}) and STL-10~\citep{coates2011analysis} ($10$ classes). Text is encoded with Qwen3-Embedding-8B and images with Qwen3-VL-Embedding-8B; both produce $4096$-dimensional embeddings, which we $\ell_2$-normalize. All embeddings are extracted once and kept frozen throughout. Since none of these datasets provides an official validation split, we hold out a stratified $10\%$ of the official training set for validation, leaving the test split untouched. Split sizes are given in Appendix~\ref{app:implementation}.

\paragraph{Baselines.}

Ideally, CAESAR would be compared with other deep clustering methods that infer the number of clusters $K$. However, such methods are rare and their code is not publicly available. Reimplementing them is not a viable option either: each one is tied to its own encoder, architecture and training pipeline, so it cannot be faithfully reproduced on our pretrained embedding space. We therefore do not include them. Instead, we compare CAESAR with two groups of methods: 

First, on both text and images, we consider methods operating on the same pretrained embedding space as CAESAR. Two baselines, like CAESAR, infer the number of clusters without access to $K$: FINCH with the hierarchy level selected by silhouette and EVoC~\citep{bot2025persistent}, a density-based method designed for embedding vectors. In contrast, a third baseline is a FINCH variant that uses the ground-truth $K$ to select, among the levels of the hierarchy, the one whose number of clusters is closest to $K$, which gives it an oracle advantage over CAESAR. Since the hierarchy only offers a few discrete levels, the selected number of clusters may differ from $K$ (Appendix~\ref{app:full_results}). Since all methods operate on exactly the same embeddings, these comparisons are controlled.

Second, on images only, we compare CAESAR with recent deep image clustering methods that are given the true number of clusters: CC~\citep{li2021contrastive}, SPICE~\citep{niu2022spice}, SCAN~\citep{van2020scan}, SIC~\citep{cai2023semantic} and TAC~\citep{li2023image}. We report their published numbers, as compiled in the benchmark of~\citet{li2023image}. These methods use a ResNet trained from scratch or frozen CLIP features and SIC and TAC also use external textual knowledge. Since the encoders differ, this comparison is not controlled. Moreover, knowing the number of clusters gives these methods a clear oracle advantage over CAESAR which must infer it. We nevertheless include them as a demanding reference point, to assess how CAESAR fares against methods that know the true number of clusters. 

Methods that use the true number of clusters, either as oracle information or as an input, are marked with $\dagger$ in the result tables.

\paragraph{Choice of the push weight.}
\label{sec:hyper}
We fix $\lambda = 0.5$ for all datasets and encoders, without dataset-specific tuning or access to the number of clusters $K$. Appendix~\ref{app:lambda} reports a sensitivity analysis over $\lambda \in \{0.1, 0.2, 0.5, 0.8\}$ on the validation splits.

\paragraph{Metrics.}
We report normalized mutual information (NMI), Adjusted Rand index (ARI;~\citealp{hubert1985comparing}) and clustering accuracy (ACC), with clusters matched to classes using the Hungarian algorithm~\citep{kuhn1955hungarian}. All metrics are computed on the test split and averaged over three seeds for methods evaluated in our experiments. The main tables report NMI and ACC; ARI is reported in Appendix~\ref{app:full_results}. Architecture, training and hyperparameter details are provided in Appendix~\ref{app:implementation}.

\subsection{Main Results}
\label{sec:exp_main}

We use NMI as our primary metric, as it naturally compares partitions with different numbers of clusters, whereas ACC relies on matching predicted clusters to ground-truth classes. In all tables, methods are ordered by increasing NMI and $\dagger$ marks methods that use the true number of clusters $K$, either as input or for model selection.

\paragraph{Text.}
Table~\ref{tab:text} compares CAESAR with standard clustering methods on AG News, all applied to the same pretrained embedding space. CAESAR outperforms every other method on both metrics, including FINCH with its level selected using the true number of clusters: it improves NMI by $11.9$ points ($61.5$ vs.\ $49.6$) and ACC by $19.8$ points ($78.3$ vs.\ $58.5$), without ever using $K$. It also recovers a number of clusters very close to the true one ($5$ vs.\ $4$; Appendix~\ref{app:full_results}). The other methods that do not use the number of clusters are far behind: EVoC reaches only $36.2$ NMI and silhouette-selected FINCH $33.6$.

\begin{table}[t]
\centering
\caption{Clustering performance on AG News (text, $\times 100$). All methods use the same pretrained embedding space. Without using the number of clusters, CAESAR outperforms all methods, including the oracle FINCH and recovers a number of clusters close to the true $K$ (Appendix~\ref{app:full_results}). raw: FINCH on the pretrained embedding space. sil.: level selected by silhouette. $K$: level whose number of clusters is closest to the true $K$. $\dagger$: oracle method, knows the true number of clusters $K$. \textbf{Bold}: best.}
\small
\setlength{\tabcolsep}{8pt}
\renewcommand{\arraystretch}{1.15}
\begin{tabular}{l c c}
\toprule
Method & NMI & ACC \\
\midrule
FINCH (raw, sil.)          & 33.6 & \phantom{0}4.5 \\
EVoC                       & 36.2 & 21.8 \\
FINCH (raw, $K$)$^\dagger$ & 49.6 & 58.5 \\
\textbf{CAESAR (ours)}     & \textbf{61.5} & \textbf{78.3} \\
\bottomrule
\end{tabular}
\label{tab:text}
\end{table}

\paragraph{Images.}
Table~\ref{tab:image} compares CAESAR with FINCH and EVoC applied to the pretrained embedding space of Qwen3-VL-Embedding, which is the direct starting point that CAESAR reorganizes and with recent deep image clustering methods. All these deep clustering methods are oracles: they are given the true number of clusters.  CAESAR uses neither. Despite this, CAESAR obtains the best NMI and ACC on CIFAR-20, the most challenging of these benchmarks, ahead of TAC and SIC. On the more separable CIFAR-10 and STL-10, the methods given $K$ keep an edge, especially in ACC, but CAESAR remains competitive in NMI and outperforms CC, SPICE and SCAN on both datasets. Among the methods applied to the same pretrained embedding space, CAESAR improves markedly over silhouette-selected FINCH on every dataset (e.g., from $29.6$ to $61.5$ NMI on CIFAR-20) and obtains a higher NMI than EVoC on all three datasets, although EVoC reaches a higher ACC on CIFAR-10 and STL-10. In NMI, CAESAR even outperforms FINCH with its level selected using the true number of clusters on all three datasets. This confirms that the reorganization, not the encoder alone, drives the results. Finally, on CIFAR-20, the silhouette criterion selects $21$ clusters for $20$ ground-truth classes, almost recovering the correct granularity without ever being given it.

\begin{table}[t]
\centering
\caption{Comparison with deep image clustering methods ($\times 100$). All baselines except EVoC and FINCH (raw, sil.) are oracle methods ($\dagger$): they know the true number of clusters $K$. Although CAESAR uses neither, it obtains the best NMI and ACC on CIFAR-20 and remains competitive in NMI on CIFAR-10 and STL-10, where it outperforms CC, SPICE and SCAN. Literature methods use ResNet or CLIP encoders and their scores are taken from the published papers~\citep{li2023image}; CAESAR, FINCH and EVoC use Qwen3-VL-Embedding features. raw: FINCH on the pretrained embedding space. sil.: level selected by silhouette. $K$: level whose number of clusters is closest to the true $K$ (Appendix~\ref{app:full_results}). \textbf{Bold}: best.}
\small
\setlength{\tabcolsep}{6pt}
\renewcommand{\arraystretch}{1.15}
\begin{tabular}{l cc cc cc}
\toprule
& \multicolumn{2}{c}{CIFAR-20} & \multicolumn{2}{c}{CIFAR-10} & \multicolumn{2}{c}{STL-10} \\
\cmidrule(lr){2-3}\cmidrule(lr){4-5}\cmidrule(lr){6-7}
Method & NMI & ACC & NMI & ACC & NMI & ACC \\
\midrule
FINCH (raw, sil.)          & 29.6 & 14.1 & 39.1 & 19.9 & 44.7 & 20.0 \\
CC$^\dagger$               & 43.1 & 42.9 & 70.5 & 79.0 & 76.4 & 85.0 \\
SPICE$^\dagger$            & 44.8 & 46.8 & 73.4 & 83.8 & 81.7 & 90.8 \\
SCAN$^\dagger$             & 48.6 & 50.7 & 79.7 & 88.3 & 69.8 & 80.9 \\
FINCH (raw, $K$)$^\dagger$ & 49.2 & 38.4 & 63.3 & 60.5 & 82.1 & 84.1 \\
EVoC                       & 56.0 & 36.0 & 71.5 & 76.0 & 82.6 & 82.4 \\
SIC$^\dagger$              & 59.3 & 58.3 & \textbf{84.7} & \textbf{92.6} & 95.3 & 98.1 \\
TAC$^\dagger$              & 61.1 & 60.7 & 83.3 & 91.9 & \textbf{95.5} & \textbf{98.2} \\
\textbf{CAESAR (ours)}     & \textbf{61.5} & \textbf{60.9} & 81.5 & 57.5 & 83.0 & 66.8 \\
\bottomrule
\end{tabular}
\label{tab:image}
\end{table}

\subsection{Component Ablation}
\label{sec:ablation_components}

To isolate the contribution of each component of CAESAR, we build it up cumulatively, starting from FINCH applied to the pretrained embedding space and adding one component at a time: the reorganization network trained with a pull-only objective, then the push term computed within each mini-batch and finally the push term computed against the full memory bank, which corresponds to our complete method. As in all experiments, $\lambda = 0.5$. Table~\ref{tab:ablation} reports the results, averaged over three seeds and over all four datasets.

Training $f_\theta$ with a pull-only objective leads to a complete collapse: with no repulsive term, all points are drawn toward one another and the geometry of the space is destroyed. The average NMI drops from $36.8$ (raw FINCH) to $2.4$. Explicit repulsion is therefore not an optional refinement but a necessary condition for learning any usable structure. Adding the push term computed within each mini-batch prevents the collapse and already improves over raw FINCH ($45.6$ NMI). However, contrasting each point only against the few examples in its mini-batch provides too few repulsive comparisons to realize the full benefit of the reorganization. Replacing the in-batch push with a push against the full memory bank yields the decisive gain, raising the average NMI to $71.9$, with ARI and ACC following the same trend. Each component is therefore necessary. Contrasting every point against the entire dataset, rather than only the current mini-batch, is the main driver of CAESAR's performance.

\begin{table}[t]
\centering
\caption{Cumulative ablation of CAESAR's components ($\times 100$, mean over three seeds, averaged over CIFAR-20, CIFAR-10, STL-10 and AG News). Starting from FINCH on the pretrained embedding space, each row adds one component; the last row is our complete method.}
\small
\setlength{\tabcolsep}{8pt}
\renewcommand{\arraystretch}{1.2}
\begin{tabular}{l c c c}
\toprule
Configuration & NMI & ARI & ACC \\
\midrule
FINCH (raw, sil.)                    & 36.8 & 26.7 & 14.6 \\
\quad + $f_\theta$ (pull only)       & \phantom{0}2.4 & \phantom{0}0.7 & 15.6 \\
\quad + push (in-batch)              & 45.6 & 30.3 & 41.6 \\
\quad + memory bank (\textbf{ours})  & \textbf{71.9} & \textbf{52.9} & \textbf{65.9} \\
\bottomrule
\end{tabular}
\label{tab:ablation}
\end{table}

\subsection{Robustness to the Encoder}
\label{sec:encoder_agnostic}

Since CAESAR reorganizes a pretrained embedding space, its benefit should not depend on the encoder that produced it. We therefore rerun the full pipeline on a much smaller encoder, CLIP ViT-B/32 ($512$-dimensional, versus $4096$ for Qwen3-VL-Embedding); results are reported in Table~\ref{tab:agnostic} (Appendix~\ref{app:encoder}).

On CLIP, CAESAR lifts the NMI of silhouette-selected FINCH from $24.3$, $32.5$ and $44.8$ to $56.5$, $80.0$ and $90.1$ on CIFAR-20, CIFAR-10 and STL-10, gains comparable to those obtained with Qwen3-VL-Embedding. On STL-10, CAESAR even performs better with CLIP than with Qwen3-VL-Embedding ($90.1$ vs.\ $83.0$ NMI) and on CIFAR-10 the two are close ($80.0$ vs.\ $81.5$). The gain thus comes from the reorganization itself rather than from a particular encoder. We keep the Qwen3 models as our main encoders because they cover both text and images, allowing a uniform evaluation across modalities.

\section{Discussion and Limitations}

\paragraph{Why CAESAR is most competitive on entangled data.}
When classes are more separable (CIFAR-10, STL-10), the pretrained embedding space already exposes a clean neighborhood structure and methods given $K$ retain an edge. When classes overlap (CIFAR-20, AG News), the first-neighbor graph of the pretrained embedding space is noisy and reshaping its geometry pays off: CAESAR matches or outperforms all baselines in NMI, including those given $K$. This is the realistic setting CAESAR targets.

\paragraph{Accuracy, granularity and the role of $K$.}
On CIFAR-10 and STL-10, CAESAR is stronger in NMI than in ACC. Although the silhouette criterion selects a number of clusters close to the true one on every dataset (Appendix~\ref{app:full_results}), ACC penalizes any mismatch through its one-to-one assignment, while NMI is less sensitive to it. On STL-10, for instance, selecting $8$ clusters for $10$ classes leaves at least two classes unmatched, which caps ACC at $80\%$. Conversely, on CIFAR-20, where $21$ clusters are selected for $20$ classes, CAESAR obtains the best ACC of all methods. A better level-selection rule than the silhouette could therefore improve ACC without changing the reorganized embedding space.

\paragraph{Scope.}
CAESAR assumes a pretrained encoder that already produces semantically meaningful embeddings: it reshapes the pretrained embedding space but cannot create structure that the encoder failed to capture. When no strong pretrained encoder is available, we expect the benefit of the reorganization to be smaller.

\section{Conclusion}
\label{sec:conclusion}

We introduced CAESAR, a lightweight method that reorganizes the pretrained embedding space of a frozen LLM/VLM encoder for graph-based clustering. A reorganization network is trained with a pull--push objective over mutual nearest neighbors, computed against a memory bank of the entire dataset and the reorganized embedding space is clustered with FINCH. Unlike most competing methods, CAESAR never uses the number of clusters.

Despite this disadvantage, CAESAR outperforms every baseline on text, including those given the true number of clusters. On images, it obtains the best NMI and ACC on CIFAR-20, ahead of methods that know the number of clusters or rely on external textual knowledge and remains competitive on CIFAR-10 and STL-10, where it still outperforms several such methods in NMI. These gains hold across encoders. Our results suggest that reorganizing pretrained embeddings is a simple and effective route to clustering realistic data when the number of clusters is unknown.

\subsection*{AI use statement}

In this work, we used generative AI tools for none of the tasks with required disclosure.
We have not used generative AI tools for developing theoretical models or conceptual frameworks, proposing or refining hypotheses, designing or providing feedback on research methodology or experiments, implementing methods or interpreting results and generating synthetic data sets, formulating mathematical claims, providing ingredients for or writing proofs, translation, cleaning and reformatting datasets and qualitative or thematic data analysis are not applicable to this work. Additionally, we used generative AI tools for identifying relevant literature and drafting parts of the related work section. We have reviewed all AI-assisted work. All references identified with generative AI were checked by the authors against the original papers to verify that they exist and that the claims attributed to them are accurate and all AI-assisted text in the related work section was revised by the authors. We take responsibility for the final content of this work, including text, claims or artifacts produced with the aid of generative AI.

\subsection*{Ethics statement}
This work does not involve human subjects or the collection of new data. All experiments use publicly available benchmark datasets (AG News, CIFAR-10, CIFAR-100, STL-10 and MNIST) and publicly released pretrained encoders. Since CAESAR reorganizes embeddings produced by these encoders, it may inherit their biases and clusters discovered on sensitive data should not be interpreted without human review. We are not aware of any other ethical concern raised by this work.

\subsection*{Reproducibility statement}
The method is fully described in Section~\ref{sec:method}. All hyperparameters and training details are given in Appendix~\ref{app:implementation} and the choice of the push weight $\lambda$ in Appendix~\ref{app:lambda}. All datasets and pretrained encoders are publicly available and the data splits are described in Section~\ref{sec:exp_setup}. Full results, including ARI and the number of selected clusters, are reported in Appendix~\ref{app:full_results}. We will release our code upon publication.

\bibliography{iclr2027_conference}
\bibliographystyle{iclr2027_conference}

\appendix

\section{Implementation Details}
\label{app:implementation}

\paragraph{Embeddings.}
All embeddings are extracted once from frozen pretrained encoders (Qwen3-Embedding-8B for text, Qwen3-VL-Embedding-8B for images and CLIP ViT-B/32 for the experiments of Section~\ref{sec:encoder_agnostic}), $\ell_2$-normalized and stored on disk. They are never updated during training.

\paragraph{Datasets.}
Table~\ref{tab:datasets} summarizes the datasets and the sizes of the training, validation and test splits.

\begin{table}[ht]
\centering
\small
\caption{Datasets used in our main experiments. All embeddings are $4096$-dimensional and kept fixed.}
\label{tab:datasets}
\begin{tabular}{lllrrrr}
\toprule
Dataset & Modality & Encoder & $K$ & Train & Val & Test \\
\midrule
AG News   & Text  & Qwen3-Embedding-8B    & $4$  & $108{,}000$ & $12{,}000$ & $7{,}600$ \\
CIFAR-10  & Image & Qwen3-VL-Embedding-8B & $10$ & $45{,}000$  & $5{,}000$  & $10{,}000$ \\
CIFAR-20  & Image & Qwen3-VL-Embedding-8B & $20$ & $45{,}000$  & $5{,}000$  & $10{,}000$ \\
STL-10    & Image & Qwen3-VL-Embedding-8B & $10$ & $4{,}500$   & $500$      & $8{,}000$ \\
\bottomrule
\end{tabular}
\end{table}

\paragraph{Training.}
The reorganization network $f_\theta$ is trained with AdamW~\citep{loshchilov2017decoupled} (learning rate $3 \times 10^{-4}$, weight decay $10^{-4}$) and a batch size of $256$, for at most $600$ epochs with early stopping (patience $20$) on the validation objective $\mathcal{L}$. The memory bank is initialized with the outputs of $f_\theta$ before training and stores the most recent $\ell_2$-normalized embedding of each training point in the reorganized embedding space; after each optimization step, the rows of the points in the current batch are updated.

\paragraph{Clustering and evaluation.}
The neighborhood size $k$ is estimated as described in Section~\ref{sec:graph}. The silhouette score used to select the FINCH level is computed on the full test split. Results of the methods we run are averaged over three seeds. All experiments were run on a single NVIDIA A100 GPU.

Table~\ref{tab:hyperparams} summarizes all hyperparameters.

\begin{table}[H]
\centering
\small
\caption{Hyperparameters of CAESAR. All values are shared across datasets and encoders, except $k$, which is estimated from each dataset.}
\label{tab:hyperparams}
\begin{tabular}{ll}
\toprule
Hyperparameter & Value \\
\midrule
Architecture of $f_\theta$ & $d \rightarrow 1024 \rightarrow 512 \rightarrow 256$ \\
Activations & SELU after each hidden layer, linear output \\
Output dimension $d'$ & $256$ \\
Optimizer & AdamW, learning rate $3 \times 10^{-4}$, weight decay $10^{-4}$ \\
Batch size & $256$ \\
Epochs & at most $600$, early stopping (patience $20$) \\
Neighborhood size $k$ & estimated by maximum curvature (Section~\ref{sec:graph}) \\
Push weight $\lambda$ & $0.5$ for all datasets (Appendix~\ref{app:lambda}) \\
Seeds & $3$ \\
\bottomrule
\end{tabular}
\end{table}

\section{Full Results}
\label{app:full_results}

Table~\ref{tab:text_ari} extends Table~\ref{tab:text} with the adjusted Rand index (ARI), Table~\ref{tab:image_ari} reports the ARI of CAESAR on the image datasets and Table~\ref{tab:n_clusters} compares the number of clusters selected by CAESAR with the true number of classes.

\begin{table}[H]
\centering
\caption{Clustering performance on AG News (text, $\times 100$), ordered by increasing NMI.
$\dagger$: given the true number of clusters. \textbf{Bold}: best in each column.}
\small
\setlength{\tabcolsep}{8pt}
\renewcommand{\arraystretch}{1.15}
\begin{tabular}{l c c c}
\toprule
Method & NMI & ACC & ARI \\
\midrule
FINCH (raw, sil.)          & 33.6 & \phantom{0}4.5 & \phantom{0}7.6 \\
EVoC                       & 36.2 & 21.8 & 30.6 \\
FINCH (raw, $K$)$^\dagger$ & 49.6 & 58.5 & 42.2 \\
\textbf{CAESAR (ours)}     & \textbf{61.5} & \textbf{78.3} & \textbf{58.3} \\
\bottomrule
\end{tabular}
\label{tab:text_ari}
\end{table}

\begin{table}[H]
\centering
\caption{ARI of CAESAR ($\times 100$) on the three image datasets: CIFAR-20 (the $20$ superclasses of CIFAR-100), CIFAR-10 and STL-10, all encoded with Qwen3-VL-Embedding-8B.}
\small
\setlength{\tabcolsep}{8pt}
\renewcommand{\arraystretch}{1.15}
\begin{tabular}{l c c c}
\toprule
Method & CIFAR-20 & CIFAR-10 & STL-10 \\
\midrule
CAESAR (ours) & 38.7 & 54.4 & 60.3 \\
\bottomrule
\end{tabular}
\label{tab:image_ari}
\end{table}

\begin{table}[H]
\centering
\caption{Number of clusters selected by the silhouette criterion on the FINCH hierarchy of CAESAR, compared with the true number of classes $K$. The number of clusters is never given to the method.}
\small
\setlength{\tabcolsep}{8pt}
\renewcommand{\arraystretch}{1.15}
\begin{tabular}{l c c c c}
\toprule
 & AG News & CIFAR-20 & CIFAR-10 & STL-10 \\
\midrule
True $K$           & $4$  & $20$ & $10$ & $10$ \\
Selected clusters  & $5$  & $21$ & $9$  & $8$ \\
\bottomrule
\end{tabular}
\label{tab:n_clusters}
\end{table}

\section{Effect of the Push Weight}
\label{app:lambda}

Table~\ref{tab:lambda-sweep} shows the effect of the push weight $\lambda$ on clustering quality, measured on validation splits. In addition to AG News, CIFAR-10 and STL-10, the sweep includes MNIST~\citep{lecun1998gradient}, which we do not report in the main evaluation because it is not covered by the benchmark of deep clustering results we compare against~\citep{li2023image}. A small weight under-penalizes non-neighbors and leaves clusters insufficiently separated, while a large weight spreads points too much. We therefore use the intermediate value $\lambda = 0.5$ for all datasets.

\begin{table}[H]
\centering
\small
\caption{Effect of the push weight $\lambda$ (NMI $\times 100$ on validation splits, averaged over AG News, CIFAR-10, STL-10 and MNIST).}
\label{tab:lambda-sweep}
\begin{tabular}{lcccc}
\toprule
$\lambda$ & 0.1 & 0.2 & 0.5 & 0.8 \\
\midrule
NMI & 41.4 & 44.5 & \textbf{68.8} & 62.0 \\
\bottomrule
\end{tabular}
\end{table}

\section{Robustness to the Encoder}
\label{app:encoder}

\begin{table}[H]
\centering
\caption{Robustness to the encoder (NMI $\times 100$). With both a large ($4096$-d Qwen3-VL-Embedding) and a much smaller ($512$-d CLIP) encoder, CAESAR yields a large improvement over silhouette-selected FINCH on the pretrained embedding space, confirming that the gain comes from the reorganization rather than from the encoder.}
\small
\setlength{\tabcolsep}{7pt}
\renewcommand{\arraystretch}{1.15}
\begin{tabular}{l l c c c}
\toprule
Encoder & Method & CIFAR-20 & CIFAR-10 & STL-10 \\
\midrule
\multirow{2}{*}{Qwen3-VL-Embedding} & FINCH (raw, sil.) & 29.6 & 39.1 & 44.7 \\
                                    & \textbf{CAESAR (ours)} & \textbf{61.5} & \textbf{81.5} & \textbf{83.0} \\
\midrule
\multirow{2}{*}{CLIP ViT-B/32}      & FINCH (raw, sil.) & 24.3 & 32.5 & 44.8 \\
                                    & \textbf{CAESAR (ours)} & \textbf{56.5} & \textbf{80.0} & \textbf{90.1} \\
\bottomrule
\end{tabular}
\label{tab:agnostic}
\end{table}

\end{document}